\documentclass{article}

\usepackage{url}
\usepackage{xcolor,colortbl}
\usepackage{amsmath,amsfonts,amssymb,mathrsfs}
\usepackage{verbatim}
\usepackage{mathrsfs}
\let\chapter\section
\usepackage[ruled,linesnumbered, noend]{algorithm2e}
\usepackage{wrapfig}
\usepackage{graphicx} 
\usepackage{xspace}
\usepackage{subfig}
\usepackage{times}
\usepackage{multirow,multicol}
\usepackage[bookmarks=true]{hyperref}
\usepackage{color}

\usepackage[final]{corl_2019} 

\title{Scene-level Pose Estimation for\\ Multiple Instances of Densely Packed Objects}

\author{
  Chaitanya Mitash, Bowen Wen, Kostas Bekris, and Abdeslam Boularias\\
  Department of Computer Science\\
  Rutgers University\\
  \texttt{\{cm1074, bw344, kb572, ab1544\}@cs.rutgers.edu} \\
}

\newcommand{\cnn}{{\tt CNN}}
\newcommand{\gan}{{\tt GAN}}

\newcommand{\trans}{T}

\newcommand{\segnet}{P_l}

\newcommand\LineSpace{\\[\dimexpr-\normalbaselineskip+0.3em]}

{\end{list}}

\begin{document}
\maketitle


\begin{abstract}
This paper introduces key machine learning operations that allow the
realization of robust, joint 6D pose estimation of multiple instances
of objects either densely packed or in unstructured piles from RGB-D
data. The first objective is to learn semantic and instance-boundary
detectors without manual labeling. An adversarial training framework
in conjunction with physics-based simulation is used to achieve
detectors that behave similarly in synthetic and real data. Given the
stochastic output of such detectors, candidates for object poses are
sampled. The second objective is to automatically learn a single score
for each pose candidate that represents its quality in terms of
explaining the entire scene via a gradient boosted tree. The proposed
method uses features derived from surface and boundary alignment
between the observed scene and the object model placed at hypothesized
poses. Scene-level, multi-instance pose estimation is then achieved by
an integer linear programming process that selects hypotheses that
maximize the sum of the learned individual scores, while respecting
constraints, such as avoiding collisions.  To evaluate this
method, a dataset of densely packed objects with
challenging setups for state-of-the-art approaches is
collected. Experiments on this dataset and a public one show that
the method significantly outperforms alternatives in terms of 6D
pose accuracy while trained only with synthetic datasets.

\end{abstract}


\section{Introduction}
\label{sec:intro}
Robot manipulation pipelines, such as in bin-picking, often integrate
perception with planning~\cite{Correll:2016aa, schwarz2018fast,
zeng2017multi}. Some systems directly compute picks without computing
the pose of objects by semantic segmentation or directly learning
grasp affordances~\cite{Gualtieri:2017aa, Mahler:2017aa,
morrison2018cartman, zeng2018robotic}. While pose-agnostic techniques
are promising, in many tasks it is important to first compute the 6D
pose of observed objects to achieve purposeful manipulation and
placement, such as in the context of packing~\cite{schwarz2018fast,
shome2019towards, Fan:2019aa}.

Estimating 6D object poses has been approached in various
ways~\cite{hodan2018bop}, such as matching of locally-defined
features \cite{aldoma2012tutorial}, or of pre-defined templates of
object models \cite{hinterstoisser2012model}, or via voting in the
local object frame using oriented point-pair
features \cite{drost2010model, Vidal:2018aa}. Most methods were
developed and evaluated for setups where each object appears once and
for relatively sparsely placed objects on tabletops.  Pose estimation
for multiple instances of the same object type and where objects may
be densely packed or in highly unstructured but dense piles has
received less attention despite its significance in application
domains, such as logistics. This is partly due to the increased difficulty
of such setups.

\begin{figure}[t]
  \centering
  \includegraphics[width=\linewidth, keepaspectratio]{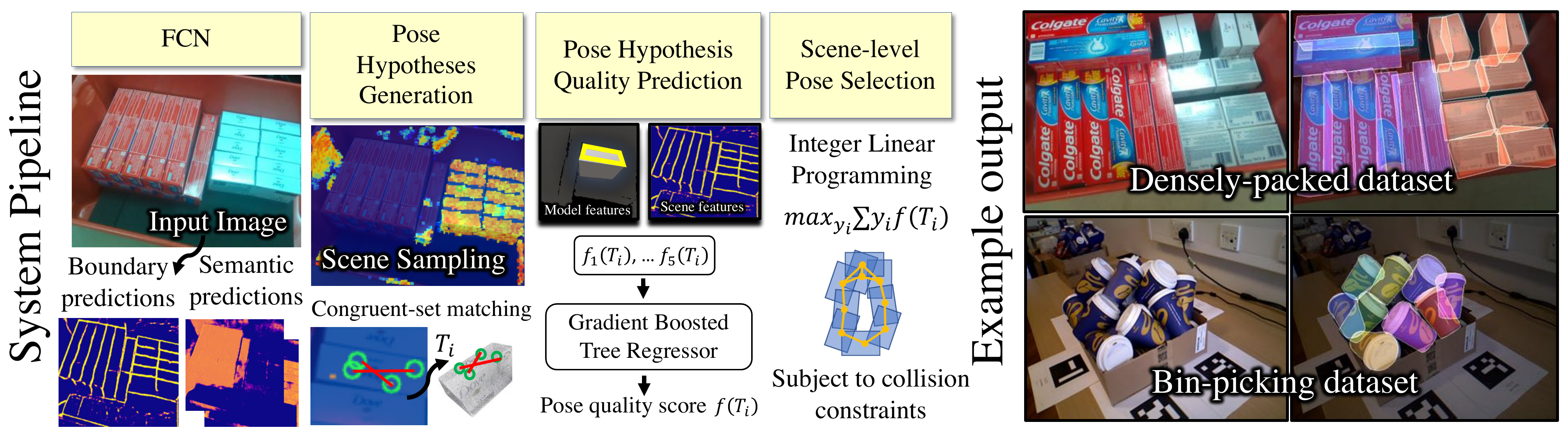}
  \vspace{-0.28in}
  \caption{\small System pipeline and example output of the proposed
  approach on densely-packed scenes}
  \label{fig:introduction}
  \vspace{-.2in}
\end{figure}

There is prior work~\cite{hodan2018bop} that provides a public
dataset~\cite{doumanoglou2016recovering} with a considerable number of
instances for the same object type. Nevertheless, it measures the
recall for estimating pose of any object instance in the
scene. This may be sufficient in certain tasks but it is a weaker
requirement than identifying the 6D pose of most, if not all, object
instances. Achieving scene-level pose estimation allows a robot to
internally simulate the world, reason about the order with which
objects can be manipulated, as well as their physical interactions and
the stability of their configuration. Scene-level reasoning can also infer missing information by considering occlusions and physical
interactions between objects.

The current work aims to improve the robustness of pose estimation
 for real-world applications, where object types appear
multiple times, and in challenging, dense configurations, such as
those illustrated in Figure~\ref{fig:introduction}, while allowing for
scene-level reasoning. It aims to do so by proposing effective machine
learning operations that depend less on manual data labeling and less
on handcrafted combinations of multiple accuracy criteria into a single objective function. This allows the true
automation of robot manipulation pipelines and brings the hope of
wider-scale, real-world deployment.

{\bf Problem Setup:} The considered framework receives as input: a) an
{\tt RGB} image $x$ and a depth map of the scene; b) a set of mesh models
$\{M_i\}_{i=1}^{K}$, one for each object type $i \in [1,K]$ present in
the scene, and c) a set $\{N_i\}_{i=1}^{K}$ expressing an upper-bound
on the number of instances for each object type $i$. The output is
object poses as a set of rigid-body transformations
$\{\{\trans^*_{ij} \}_{j=1}^{N_i} \}_{i=1}^{K}$, where each $T^*_{ij}
= (t^*_{ij}, R^*_{ij})$ captures the translation $t^*_{ij} \in R^3$
and rotation $R^*_{ij} \in SO(3)$ of the $j^{th}$ instance of object
type $i$ in the camera's reference frame, for each of the $K$ object
types present in the scene.

Figure~\ref{fig:introduction} summarizes the considered pipeline: a)
{\tt CNN}s are used to detect semantic object classes and visible
boundaries of individual instances, b) then, a large set of candidate
6D pose hypotheses are generated for each object class, c) quality
scores are computed for each hypothesis, and d) scene-level reasoning
identifies consistent poses that maximize the sum of individual
scores. In the context of this pipeline, the contribution of this work
relative to state-of-the-art methods is two-fold:

\noindent {\bf A. Adversarial training with synthetic data for robust
object class and boundary prediction:} Machine learning approaches
have become popular in pose estimation, both in end-to-end
learning \cite{xiang2017posecnn, kehl2017ssd} and as a pipeline
component \cite{brachmann2014learning, mitash2018robust}. They
require, however, large amounts of labeled data. Recent approaches aim
to solve single-instance pose estimation by training entirely in
simulation \cite{Tremblay:2018aa, mitash2018robust,
Sundermeyer:2018aa}. The proposed method also utilizes labeled data
generated exclusively in simulation to train a {\tt CNN} for semantic
segmentation. Nevertheless, {\tt CNN}s are sensitive to the domain gap
between synthetic and real data. The proposed training aims to mimic
the physics of real-world test scenes and bridges the domain gap by
using a generative adversarial network, as explained in
Section~\ref{sec:learning}. A key insight to improve robustness in the
multi-instance case is that the network is simultaneously trained to
predict object visibility boundaries. A thesis of this work is that
boundaries learned on {\tt RGB} images are more effective than boundaries
detected on depth-maps \cite{Vidal:2018aa} to guide and constrain the
search for 6D poses, especially in tightly-packed setups.


{\bf B. Scene-level reasoning by automatically learning to evaluate
pose candidate quality:} This work finds the best
physically-consistent set of poses among multiple candidates by
formulating a constraint optimization problem and applying an {\tt
ILP} solver as shown in Section~\ref{sec:selection}. The objective is
to select pose hypotheses that maximize the sum of their individual
scores, while respecting constraints, such as avoiding perceived
collisions. Scene-level optimization has been previously approached as
maximizing the weighted sum of various geometric
features~\cite{aldoma2012global}. The weights characterizing the
objective function, however, were carefully handcrafted. This work
shows it is possible to learn the distance of a given candidate pose
from a ground-truth one by using a set of various objective functions
as features. A gradient boosted tree~\cite{elith2008working} is
trained to automatically integrate these objectives and regress the
distance to the closest ground truth pose. The objectives indicate how
well a candidate hypothesis explain the predicted object segments, the
predicted boundaries, the observed depth and local surface normals in
the input data. Prior related work has used tree
search~\cite{Narayanan:2016aa, sui2017goal, mitash2018improving} to
reconstruct the scene by sequentially placing objects. These prior
approaches, however, were restricted to a small number of objects or
fewer degrees-of-freedom due to computational overhead. The proposed
{\tt ILP} solution is quite fast in practice and scales to a large
number of objects. For the images of Figure~\ref{fig:introduction},
the scene-level optimization is achieved in a few milliseconds.


\section{Semantic and Boundary Predictions}
\label{sec:learning}
\vspace{-.1in}

Fully Convolutional Networks ({\tt FCN}s) \cite{long2015fully,
noh2015learning} are popular semantic segmentation tools. They have
also been used for object contour detection~\cite{yang2016object} and
predicting multiple instances of an object type~\cite{li2017fully,
kirillov2017instancecut}. These networks are increasingly being
trained in simulation \cite{mitash2017self, tobin2017domain,
hinterstoisser2018pre, Tremblay:2018aa} to alleviate the need for
large amounts of labeled data. The domain gap between the data
generated in simulation and real data can lead to noisy predictions
and greatly affect pose estimation accuracy. Several recent methods
have been developed to bridge this domain gap~\cite{CycleGAN2017,
Tsai_adaptseg_2018}. The current work subscribes to these ideas and:
a) exploits the constraints available in robotic setups to simulate
scenes with realistic poses, while b) uses adversarial training with
unlabeled real images to bridge the gap between the labels predicted
in synthetic data with those predicted in real ones.

This work proposes the use of a {\tt CNN} to predict per-pixel
semantic classification and a classification of whether a pixel is a
visible object boundary. The data for training the {\tt CNN} are
generated in simulation with a physics engine and a renderer. The
simulation samples a bin pose and a camera pose given the robot's
workspace. Each scene is created by randomly sampling, within a
pre-specified domain, the number and 6D poses of objects, the color of
the bin, and the placement and intensity of the illumination
sources. Finally, the scene is rendered to obtain a color image, a
depth map, per-pixel class labels and visible instance boundary
labels. The simulation generates a wide range of training data for
domain randomization and robustness to domain gap issues.
Nevertheless, domain gap still exists between synthetic data and data
acquired through real sensors as it is hard to capture the domain of
object material's interactions with various illumination sources in
the environment.

\vspace{-0.1in}
\begin{figure}[ht!]
  \centering
  \includegraphics[width=\linewidth, keepaspectratio]{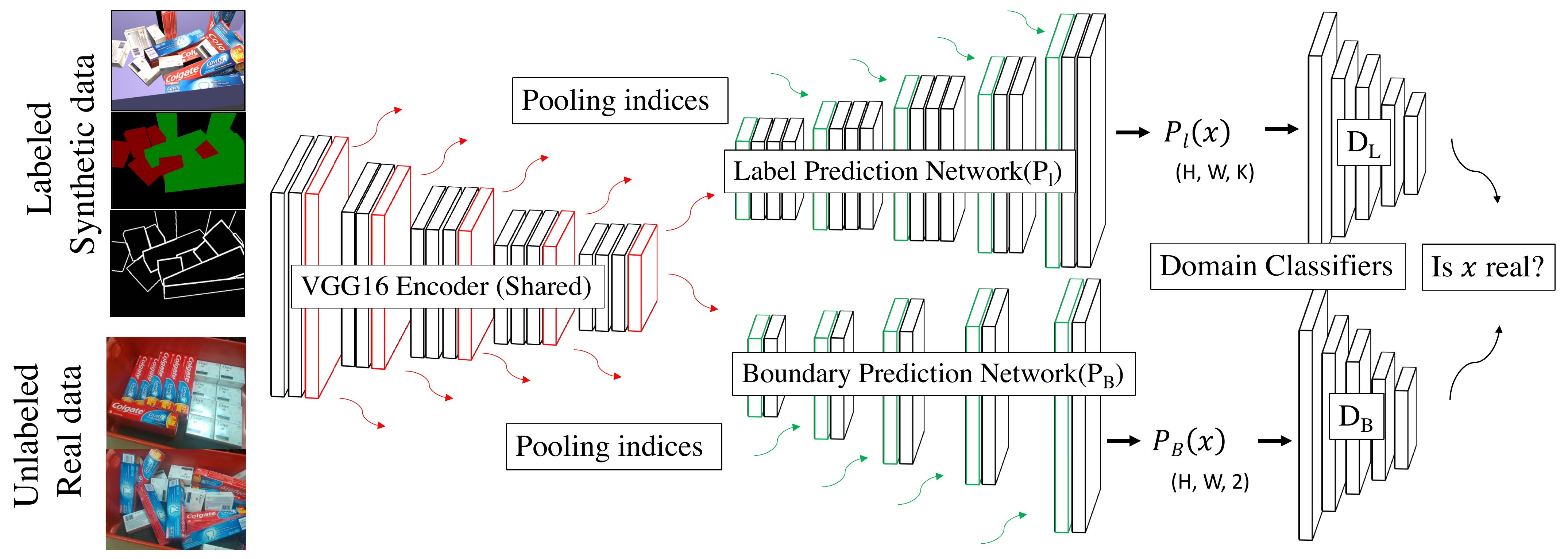}
\vspace{-0.15in}
\caption{\small {\tt CNN} architecture for semantic classes and boundary prediction.}
\label{fig:cnn}
\vspace{-0.15in}
\end{figure}

The generative adversarial network (\gan), shown in
Figure~\ref{fig:cnn}, performs the semantic and boundary detection
tasks, while also adapting the output predictions on unlabeled real
images to resemble the predictions on synthetic images. It consists of
a shared {\tt VGG16} encoder that stacks five blocks of convolution,
batch normalization and max pooling layers. The network branches out
into two decoders for semantic and boundary classifications. These are
fully convolutional decoders with unpooling indices passed from
corresponding max pooling blocks in the encoder section. The outputs
of both decoders are passed to a corresponding fully convolutional
discriminator network.

The network is trained by taking as input a synthetic image $x_s$ and its ground-truth label $Y(x_s)$. It also receives as input unlabeled real image $x_r$. The output $\segnet(x_s)$ of the label prediction network on image $x_s$ (or corresponding output on $x_r$) is then passed on to the label discriminator $D_L$ whose task is to classify correctly if the prediction is on real or on synthetic data. Along with the objective of correctly labeling the synthetic image $x_s$, the label prediction network should also confuse $D_L$ into classifying
$\segnet(x_r)$ as an output of a sample coming from the synthetic domain. The objective of the semantic labeling network is defined as:




\vspace{-0.2in}
\begin{eqnarray*}
\mathcal L_{sem} = - \sum_{h,w}\sum_{k \in K} Y(x_s)_{(h,w,k)} \log P_l(x_s)_{(h,w,k)} - \lambda_a\sum_{h, w}\log D_L(P_l(x_r))_{(h,w,0)}
\end{eqnarray*}
\vspace{-0.2in}

\noindent where $P_l(x_s)$ is the per-pixel K-channel (for K object classes) output from the labeling network, $(h,w)$ are pixel coordinates, and $\lambda_a$ is a weight factor. 
$D_L(P_l(x))_{(h,w,0)}$ is the predicted score of $x$ in pixel $(h,w)$ of being a synthetic image.  
The domain classifier's objective is specified as:

\vspace{-0.15in}
\begin{eqnarray*}
\mathcal L_{D_L} = - \sum_{h, w} \log D_L(P(x_s))_{(h, w, 0)} - \sum_{h, w} \log D_L(P(x_r))_{(h, w, 1)}
\end{eqnarray*}
\vspace{-0.15in}

A similar \gan~objective is used to simultaneously train the boundary predictor and discriminator.

\section{Scene-level Pose Selection}
\label{sec:selection}
\vspace{-.1in}

This section formulates the scene-level objective and presents how to
address it in a computationally efficient manner, despite its
computational hardness. Given the semantic and boundary predictions, a
set of 6D pose hypotheses is generated,
$\{\{\trans_{ij} \}_{j=1}^{H_i} \}_{i=1}^{K}$, where $H_i$ denotes the
number of hypotheses for each object class ($H_i \gg N_i$).  Pose
candidates can be generated via learning \cite{wang2019densefusion},
{\tt RANSAC}~\cite{brachmann2014learning} hough
voting~\cite{drost2010model}, or by incorporating distinctive
geometric features~\cite{mitash2018robust}.

The current work uses a stochastic representation of output from
segmentation for hypotheses generation via congruent set
matching \cite{mellado2014super,mitash2018robust}. It iteratively
samples a set of points (called a {\it base}) from the observed point
cloud such that the points in each set belong with high probability to
a single object instance. The sampled point sets are then matched to
congruent sets on corresponding object model to generate candidate
rigid transformations. A key feature of our hypotheses generation process, compared to~\cite{mellado2014super,mitash2018robust}, is that the boundary
predictions from the previous step are utilized to limit the selection
of points within a single object instance. An adaptive sampling
process is used to cover all instances of the same object by enforcing
dispersion. The detailed formulation for  the hypotheses generation process can be found in
Appendix A.

The final objective is to select for an object category $i$, a subset
of poses
$\{\trans^*_{ij} \}_{j=1}^{N_i} \subset \{\trans_{ij} \}_{j=1}^{H_i}$,
which maximizes the total sum of scores $f(\trans^*_{ij})$ while
avoiding collisions between objects when assigned to those poses.  In
other terms, the intersection between the volume occupied by any two
object instances in the scene should be empty. The approach first
identifies a set that contains all pairs of poses, which conflict with
each other. This set is defined as $\mathcal C = \{ \{(i,j),
(i',j')\} \text{ s.t. } | \mathcal V(M_i, \trans_{ij}) \cap \mathcal
V(M_i, \trans_{i'j'})| > \epsilon_{v} \}$ where $i,i'\in\{1,..,K\},
j\in\{1,.., H_i\}, j'\in\{1,.., H_i'\}$ and $\mathcal V(M, \trans)$ is
the volume occupied by a model $M$ when placed according to pose
$\trans$. $\epsilon_{v}$ is the maximum volume of tolerated collisions
between objects in the scene. This positive error term is necessary
because the best poses among the sampled ones may induce slight
collisions between objects, which can often be corrected afterwards.

Pose selection is formulated as a constraint optimization problem,
which can be solved by {\tt ILP}. A set of binary variables
$\{\{y_{ij} \}_{j=1}^{H_i} \}_{i=1}^{K}$ is defined where
$y_{ij} \in \{0,1\}$. Each variable $y_{ij}$ can be seen as an
indicator variable on whether pose hypothesis $\trans_{ij}$ is
included in the set of selected posed
$\{\trans^*_{ij} \}_{j=1}^{N_i}$. Then, the optimization problem is as
follows:

\vspace{-0.3in}
\begin{eqnarray*}
\underset{\{y_{ij} \}} {\textrm{max}} && \sum_{i=1}^{K} \sum_{j=1}^{H_i} y_{ij} f(\trans_{ij})
\end{eqnarray*}
\vspace{-0.3in}
\begin{eqnarray*}
\textrm{subject to:}&& \forall i\in \{1,\dots, K\}:  \sum_{j=1}^{H_i} y_{ij} \leq N_i \\
&& \forall \{(i,j), (i',j')\} \in \mathcal C : y_{ij} + y_{i'j'} \leq 1.
\end{eqnarray*}
\vspace{-0.3in}

The first constraint ensures that the number of poses selected for
each category of object $i$ do not exceed the number $N_i$ of
instances.  The second constraint ensures that poses that are
conflicting cannot be selected. This problem is equivalent to the
Maximum-Weight Independent Set Problem ({\tt MWISP}), with an
additional constraint related to the number of selected poses.  {\tt
MWISP} is NP-hard, and there are no $\frac{1}{n^{1-\epsilon}
}$-approximations for any fixed $\epsilon > 0 $ where $n$ is the
number of variables $\{ y_{ij} \}$~\cite{Hastad96cliqueis}.  In
practice, however, and for the problems considered here, an exact
solution can be found very fast (in milliseconds) using modern {\tt
ILP} solvers for scenes containing a few hundreds of candidate
poses. This is because the poses tend to cluster into cliques around
specific instances, with a small number of constraints between poses
in different cliques. Moreover, the objective function is monotone
sub-modular as it is a linear function of a subset~\cite{0001G14}.
Therefore, greedy optimization is guaranteed to find in linear time a
solution that is at least $(1-\frac{1}{e})$ fraction of the optimal.
An approximate solution can also be found in linear time with an {\tt
LP} relaxation.  After solving the {\tt ILP}, the poses
$\{\trans^*_{ij} \}_{j=1}^{N_i}$ are constructed from
$\{\trans_{ij} \}_{j=1}^{H_i}$ by keeping those for which $y_{ij} =
1$.

\vspace{-.1in}

\section{Pose Hypothesis Quality Evaluation}
\label{sec:hypeval}


\vspace{-.1in}

\begin{table}
\begin{tabular}[t]{m{0.2cm} m{12.6cm}}
\hline
\multicolumn{2}{c}{Notation}\\
\hline
&\LineSpace
\multicolumn{2}{c}{$B(T)$: Visible boundary pixels when the object model is placed at pose $T$ and rendered}\\
\multicolumn{2}{c}{$S_B$: Scene boundary pixels predicted by the \cnn.}\\
\multicolumn{2}{c}{$V(T)$: Visible portion of the the object model when placed at pose $T$ and rendered.}\\
\multicolumn{2}{c}{$\delta_S, \delta_B$: Surface and boundary matching distance thresholds.}\\
\multicolumn{2}{c}{$\mathcal D_S(p, T, O_D)$: Depth distance between rendered image and observed depth image $O_D$ at pixel $p$.}\\
\multicolumn{2}{c}{$\mathcal D_B (p, T, S_B)$: Distance between rendered boundary and predicted boundary at pixel $p$.}\\
&\LineSpace
\hline
\multicolumn{2}{c}{Model-to-Scene consistency features}\\
\hline
&\LineSpace
$f_1$ & $|B(T)\cap S_B|/|B(T)|$: fraction of pixels in model boundary that match the scene boundary.\\
&\LineSpace
\hline
&\LineSpace
$f_2$ & $|B(T)\cap S_B|/{|V(T) \cap S_B|}$: fraction of pixels in scene boundary within the visible model region that match the model boundary.\\
&\LineSpace
\hline
&\LineSpace
$f_3$ & ${\sum_{p \in V(T)} \mathbf{1}_{\mathcal D_S(p, T, O_D) < \delta_S}}/{|V(T)|}$: fraction of pixels in visible model region that is sufficiently aligned in terms of depth to the observed data.\\
&\LineSpace
\hline
\multicolumn{2}{c}{Scene alignment features}\\
\hline
&\LineSpace
$f_4$ & $\sum_{p \in V(T)} P_l(p) \mathcal S(p,T,O_D)$: surface alignment score weighted by the corresponding label probability. Similarity score given by $\mathcal S(p,T,O_D) = (1 - \frac{1}{\delta_S} \mathcal D_S(p, T, O_D))\mathcal N(p, T)$ and it considers depth distance and surface normal similarity.\\
&\LineSpace
\hline
&\LineSpace
$f_5$ & $\sum_{p \in B(T)} (1 - \frac{1}{\delta_B} \mathcal D_B (p, T, S_B))$: boundary alignment score based on distance between point on model boundary and it's nearest point in the predicted boundary set.\\
&\LineSpace
\hline
\end{tabular}
\vspace{0.02in}
\caption{\small Description of features indicating good pose alignment with sensory input}
\label{table:quality_features}
\vspace{-.25in}
\end{table}

Scene-level optimization, presented in Section~\ref{sec:selection}, requires that a single quality score $f(T)$
is assigned to each pose $T$ in the hypotheses set. The proposed score
function considers five indicators $f_{1:5}(T)$ of a good alignment
for each pose candidate $T$, shown in
Table~\ref{table:quality_features}. A straightforward solution is to
define $f$ as a weighted sum of its components. Nevertheless, the
resulting poses would heavily depend on the choice of the weights.
Choosing the right weights manually for every new object type is not
trivial. Instead, a key aspect of this work is to learn the objective
function $f(T)$ using $f_{1:5}(T)$ as features, i.e., $f(T) =
h(f_{1:5}(T))$. The function $h$ is learned by minimizing the
following loss:

\vspace{-0.18in}
\begin{eqnarray}
\underset{h} {\textrm{min}} \sum_{(T,f_{1:5}(T),T^{g}) \in \mathcal T_{train}}  \Big( h(f_{1:5}(T)) -  e_{ADI}(T,T^g) \Big)^2,
\label{GBRT_obj}
\end{eqnarray}
\vspace{-0.18in}

\noindent where $e_{ADI}(T,T^g)$ is the distance between a given pose $T$ and a ground-truth pose $T^g$. The distance $e_{ADI}(T,T^g)$ is
computed using the {\tt ADI} metric, which is frequently used in the
literature for evaluating pose
estimations~\cite{hinterstoisser2012model}. This metric is explained
in the experimental section.

\begin{figure}[h]
\vspace{-.1in}
  \centering \includegraphics[width=0.95\linewidth]{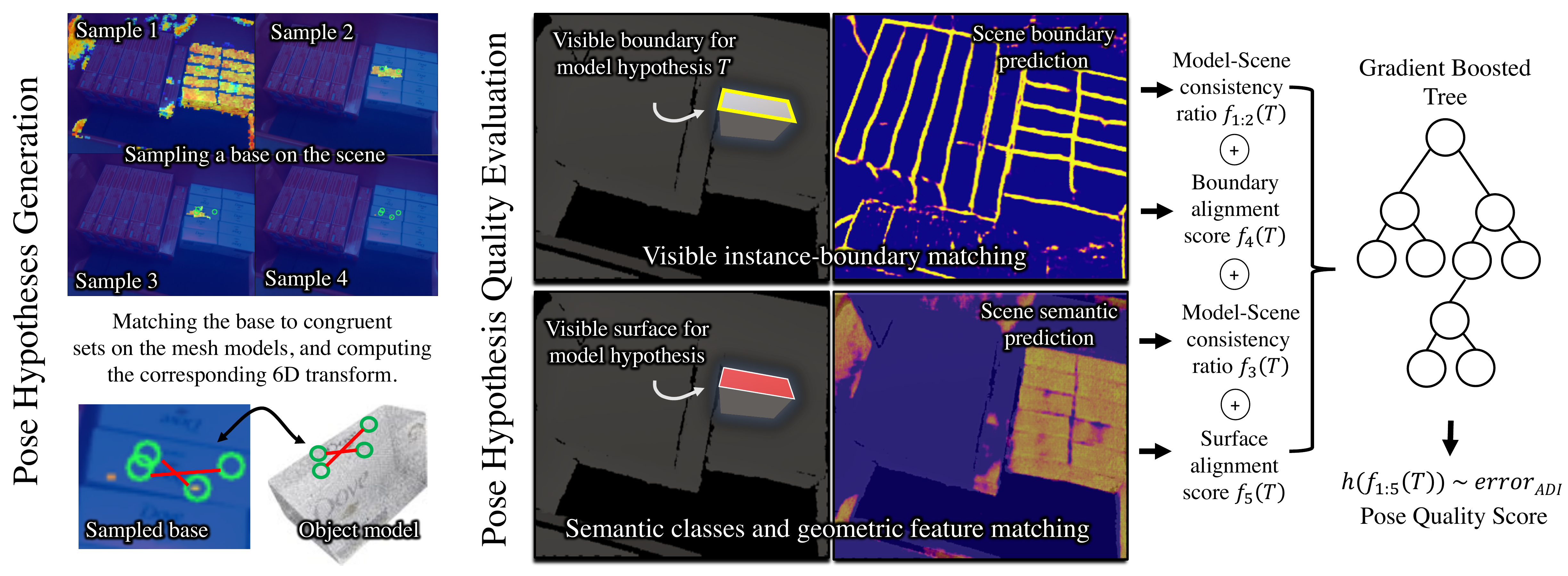}
\vspace{-0.08in}
\caption{\small Regressing the hypothesis quality given various alignment features.}
\vspace{-.1in}
  \label{fig:qualitative}
\end{figure}

The training data set $\mathcal T_{train}$ is collected by simulating
different scenes in a physics engine in the same way as described in
Section.~\ref{sec:learning}. For each scene, the {\tt CNN} predicts
semantic and object boundaries and a large number of pose hypotheses
are sampled. For each hypothesis T, their alignment features
$f_{1:5}(T)$ and corresponding scores $e_{ADI}(T,T^g)$ are computed
based on the closest ground-truth pose. The regression learning
problem is to find a function $h$ that maps features $f_{1:5}(T)$ of a
pose $T$ to its actual distance from the corresponding ground-truth
pose $T^g$.

This work adopts the Gradient Boosted Regression Trees ({\tt GBRT}) to
solve the optimization problem in Equation~\ref{GBRT_obj}. {\tt GBRT}s
are well-suited for handling variables that have heterogeneous
features~\cite{elith2008working}. {\tt GBRT}s are also a flexible
non-parametric approach that can adapt to non-smooth changes in the
regressed function using limited data, which often occurs when dealing
with objects that have different shapes and sizes. An implementation of {\tt GBRT} is available in the
Scikit-learn library~\cite{pedregosa2011scikit}.




\section{Experiments}
\label{sec:evaluation}
 \vspace{-.1in}

This section describes the experimental study performed on 2 datasets
of scenes with multiple instances of objects. In this study, the
recall for pose estimation is measured based on the error given by
{\tt ADI}~\cite{hinterstoisser2012model}, which measures distances
between poses $T_1$ and $T_2$ given an object mesh model $M$:

\vspace{-0.2in}
\begin{align*}
e_{ADI} (T_1,T_2) = \text{avg}_{b_1 \in M} \text{min}_ {b_2 \in M} || T_1(b_1,M) - T_2(b_2,M)||_2,
\end{align*}
\vspace{-0.2in}

\noindent where $T(b,M)$ corresponds to point $b$ after applying
transformation $T$ on $M$. Given a ground-truth pose $T^g$, a true
positive is a returned pose $T$ that has $e_{ADI}(T, T^g) < k_l d_l$,
where $k_l$ is a fraction set to $\frac{1}{10}$, while $d_l$ is the
diameter of the object model calculated as the maximum distance
between any two points on the model.

The two datasets include a public dataset called {\it bin-picking
dataset} \cite{doumanoglou2016recovering} and a new one developed as
part of this submission. The {\it bin-picking dataset} contains two
object types and 3 scenarios (Figure~\ref{testcases}). These scenarios
present clutter of objects with high occlusion rate. For this dataset,
segmenting objects in color images is challenging but depth cues can
be used to find the object boundaries. This leads to depth-based
approaches achieving high estimation success on this dataset without
color information.  The new dataset, henceforth called the {\it
densely-packed} dataset, comprises of $30$ unique scenes with two
object categories. Each scene contains $15$ to $19$ different
instances of these objects, which were manually labeled with 6D pose
annotations. There are two types of scenes in this dataset,
as illustrated below. In one type, denoted as
scenario 1, the instances are tightly packed next to each other. This
case is particularly challenging because the surfaces of multiple
instances are aligned, which makes it difficult to use depth
information for segmentation. The dataset and the code is shared
alongside the paper.

\begin{table}[h!]
\vspace{-.1in}
\includegraphics[width=\linewidth, keepaspectratio]{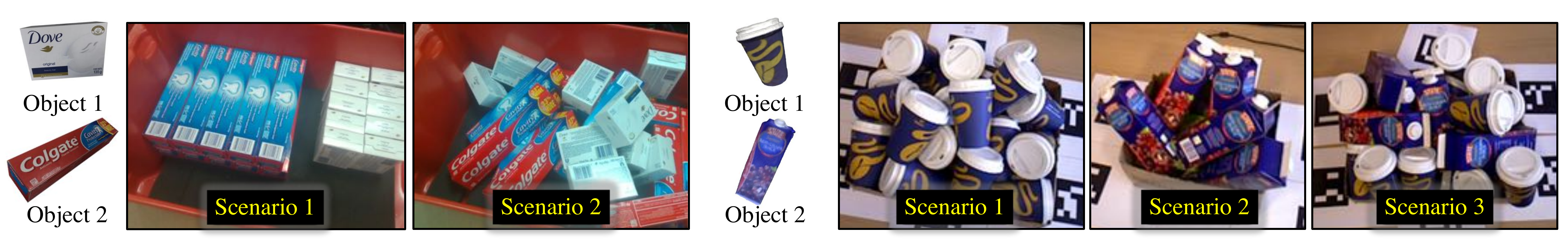}
\caption{Object type and scenarios in the (left) {\it
densely-packed} and (right) {\it bin-picking} dataset.}
\label{testcases}
\parbox{.55\linewidth}{
\begin{tabular}[t]{cccc}
\hline
Approach&o1&o2&Mean\\
\hline
OURS & \bf{79.9} & \bf{85.2} & \bf{82.1}\\
Hinterstoisser at. al \cite{hinterstoisser2012model} & 37.0 & 65.6 & 49.3\\
PPF-Voting \cite{drost2010model} & 30.1 & 57.6 & 41.9\\
Buch at. al \cite{buch2017rotational} & 11.2 & 31.7 & 19.9\\
LCHF \cite{doumanoglou2016recovering} & 16.2 & 44.3 & 28.3\\
MRCNN-StoCS \cite{he2017mask, mitash2018robust} & 42.8 & 68.3 & 53.7\\
PoseCNN$^{**}$ \cite{xiang2017posecnn} & 15.0 & 46.9 & 28.7\\
PoseCNN + ICP$^{**}$ \cite{xiang2017posecnn} & 56.8 & 80.6 & 67.0\\
DOPE$^{**}$ \cite{Tremblay:2018aa} & 51.0 & 70.6 & 60.8\\
DOPE \cite{Tremblay:2018aa} & 3.5 & 10.5 & 6.5\\
\hline
\end{tabular}
\caption{Pose retrieval recall rate on {\it densely-packed} dataset. $^{**}$ \cite{xiang2017posecnn, Tremblay:2018aa} tested on synthetic version.}
\label{table:other_methods}
}
\hfill
\parbox{.42\linewidth}{
\begin{tabular}[t]{ccc}
\hline
Approach&o1&o2\\
\hline
OURS & \bf{64.1} & \bf{55.7}\\
Buch at. al \cite{buch2017rotational} & 63.8 & 44.9\\
PPF-Voting \cite{drost2010model} & 47.4 & 27.9\\
LCHF \cite{doumanoglou2016recovering} & 33.5 & 25.1\\
Tejani et. al. \cite{Tejani:2014aa} & 31.4 & 24.8\\
\hline
\end{tabular}
\caption{Recall on {\it bin-picking dataset}. Results for other
approaches are obtained from \cite{buch2017rotational}. Several object
instances were completely hidden. This leads to lower recall rates even when the algorithms could retrieve all the visible instances.}
\label{table:bin_picking}
}
\vspace{-.2in}
\end{table}

\subsection{Evaluation against recent pose estimation techniques}
\vspace{-.1in}

Several state-of-the-art pose estimation techniques are evaluated on
the above datasets (Table ~\ref{table:other_methods} and
Table ~\ref{table:bin_picking}). A popular template-matching work
\cite{hinterstoisser2012model} matches templates extracted from {\tt RGB}
and depth rendering of {\tt CAD} models. It fails on several occasions
as the templates are not robust to occlusion and varying lighting
conditions. Approaches based on hough voting with point-pair features
\cite{drost2010model, buch2017rotational} achieve
high success on the {\it bin-picking} dataset but fail to do so on the
{\it densely-packed} dataset. These approaches detect multiple object
instances by considering the peaks in hough voting space, several of
which are false positives by virtue of aligned surfaces in the packed
boxes scenario. {\tt LCHF} \cite{doumanoglou2016recovering} is
tailored to handle multiple instances of the same object
category. Even after carefully tuning the weights of the optimization
function and relaxing the criteria for the number of pose candidates
to select, the recall from this approach is rather low. This can be
majorly attributed to differences in pre-defined template descriptors
between the scene and the object model used for matching local
patches. {\tt LCHF} \cite{doumanoglou2016recovering} also includes an
active vision component not considered in this evaluation.

Next, recent deep-learning based techniques for pose estimation are
evaluated. {\tt StoCS}~\cite{mitash2018robust} matches point cloud
object models to stochastic output from a {\tt CNN} trained for
semantic segmentation with synthetic data. One way to apply {\tt
StoCS} for multi-instance estimation is to integrate it with an
instance segmentation technique, such as {\tt Mask
R-CNN}~\cite{he2017mask}. {\tt Mask R-CNN} was trained with synthetic
data and used to extract top $10$ instances of each object type
according to the detection probability. Pose estimation was performed
using {\tt StoCS} for each of these individual segments. There are two
major limitations of this combination. The first is the use of
deterministic instance boundaries leading to segmentation noise that
cannot be recovered during pose estimation. The second is that
visibility is not considered when matching the model to the detected
segment, which can lead to several incorrect poses achieving high
alignment scores.

{\tt PoseCNN}~\cite{xiang2017posecnn} is an end-to-end learning
approach. It includes a network branch for semantic
segmentation. Pixels belonging to an object class then vote for the
object centroid's location. Based on the peaks in voting, the center
is localized and corresponding inliers are used to find a region of
interest ({\tt RoI}). Features of the {\tt RoI} regress in a separate
branch of the network to output the object's rotation. {\tt PoseCNN}
was originally developed for single instances but via non-maximal
suppression over the output of hough voting, it can be adapted for
multiple instances. To eliminate domain gap from the scope of testing,
{\tt PoseCNN} was tested on a synthetic dataset with the same
scenarios as the ones in real testing. {\tt PoseCNN} outputs an object
pose that could be used as an initialization for a depth-based {\tt
ICP}-like process that utilizes perturbations and local search for
refinement. Overall, object symmetry and tight-packing scenarios make
the simultaneous training of the multiple network branches hard to
converge and the final {\tt ICP} process less effective in these
scenarios.

{\tt DOPE}~\cite{Tremblay:2018aa} is another learning-based approach
that recovers 6D pose via perspective-n-point ({\tt PnP}) from
predictions of 3D bounding-box vertices projected on the image. It
aims to bridge the simulation-to-reality gap by a combination of
domain randomization and photo-realistic rendering. The open-sourced
rendering engine {\tt NDDS}~\cite{to2018ndds} used in {\tt DOPE},
however, does not provide access to photo-realistic rendering. Thus,
the comparisons were made against the {\tt DOPE-DR} version. The
neural network output is composed of a {\tt belief map} used to find
the projected vertices by a local peak search as well as an {\it
affinity map}, which indicates the direction from projected vertices
to their corresponding center for assignment. Nevertheless, when
multiple instances of the same object are placed next to each other,
some 2D vertices significantly overlap around the border of two
neighboring instances. This makes the assignment of vertices to the
correct center problematic and degrade the performance of ({\tt PnP}),
since it requires relatively precise 2D-3D point correspondences. {\tt
DOPE} uses only color information without depth, which is a
disadvantage when compared to other methods in
Table~\ref{table:other_methods}. {\tt DOPE} was trained from scratch
with synthetic data generated by following the same pipeline as
presented in~\cite{Tremblay:2018aa}. Using the best tuned parameters
for domain randomization and post-processing steps, the pose
estimation recall was measured on a synthetic validation set and on
the real test set.
\vspace{-.1in}

\subsection{Ablation study of the proposed technique}
\vspace{-.1in}

The \cnn~in Section~\ref{sec:learning} is trained with 20,000 scenes
generated in simulation by randomly dropping objects in the bin. It is
then fine-tuned with images rendered from $50$ simulated scenes of
tightly-packed objects. The networks were trained using the Adam
optimizer with an initial learning rate of $2\times 10^{-3}$. The
weight $\lambda_a$ for the {\tt GAN} loss is set to $10^{-3}$. To
handle the class imbalance in the boundary network, the ratio of
boundary to non-boundary pixels was computed in every iteration of the
training and used to weight the respective loss terms.

\begin{table}[h]
\parbox{.45\linewidth}{
\begin{tabular}[t]{cccc}
\hline
Training&o1&o2&All\\
\hline
Adapt-finetune (ours) & {\bf 79.9} & 85.2 & {\bf 82.1}\\
Adapt-nofinetune & 74.5 & 84.6 & 78.8\\
Cyclegan \cite{CycleGAN2017} & 75.6 & {\bf 88.2} & 81.0\\
No adaptation (rgb) & 69.8 & 79.5 & 73.9\\
\hline
\end{tabular}
\vspace{0.05in}
\caption{\small Evaluating different training strategies.}
\label{train_ablation}
}
\hfill
\parbox{.5\linewidth}{
\begin{tabular}[t]{cccc}
\hline
ILP Objective function &o1&o2&All\\
\hline
Optimal & 87.6 & 90.8 & 88.9\\
Learned & 79.9 & 85.2 & 82.1\\
Manual (scene + model) & 76.4 & 83.1 & 79.2\\
Manual (scene) & 59.8 & 82.6 & 69.6\\
\hline
\end{tabular}
\vspace{0.05in}
\caption{\small learned vs manually-defined objective.}
\label{obj_function}
}
\parbox{.45\linewidth}{
\begin{tabular}{cccc}
\hline
Method& Selected & All-Candidates\\
\hline
{with boundary} & 82.1 & 92.9\\
{w/o boundary} & 58.7 & 78.0\\
\hline
\end{tabular}
\vspace{0.05in}
\caption{\small Effect of using boundaries for hypotheses generation.}
\label{hg_boundary}
}
\hfill
\parbox{.52\linewidth}{
\includegraphics[width=\linewidth]{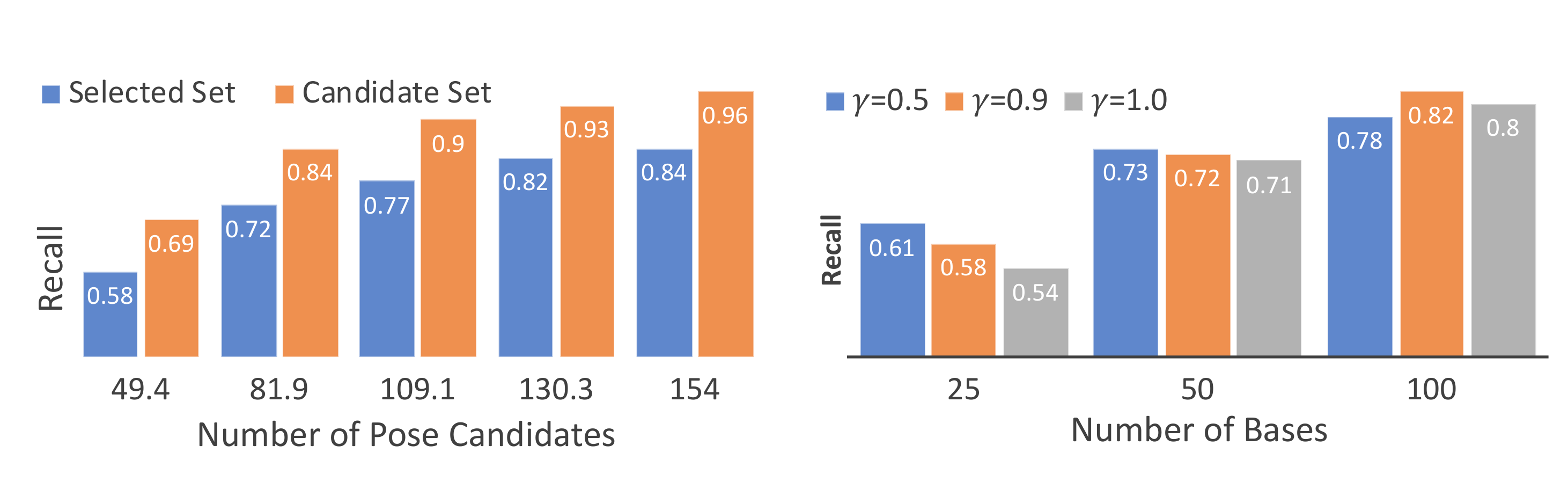}
\caption{\small Recall as a function of number of pose candidates and the dispersion parameter ($\gamma$)}
\label{hg_ablation}
}
\end{table}

Table~\ref{train_ablation} indicates that label-space adaptation is
the most effective training strategy in this case and even more so
when scenes that mimic the packing scenario were used. Training solely
on synthetic data with no adaptation significantly reduces the
performance.  An unpaired image translation approach, {\tt
Cycle-GAN} \cite{CycleGAN2017}, achieves comparable performance. But
with no semantic constraints, it biases the transfer for dominant
classes in the dataset (o2 in this case), and also cannot deal with
background clutter. Figure~\ref{fig:qualitative} shows synthetic
training images for different datasets and boundary predictions on
real images of corresponding datasets. To evaluate the generalization
capacity of the training process, the network was also trained for the
Occluded-Linemod dataset \cite{brachmann2014learning} that contains 8
object classes and unseen background clutter. Even then, the training
was able to predict boundaries of only concerned objects.


\begin{figure}[h]
\vspace{-0.2in}
\includegraphics[width=\linewidth, keepaspectratio]{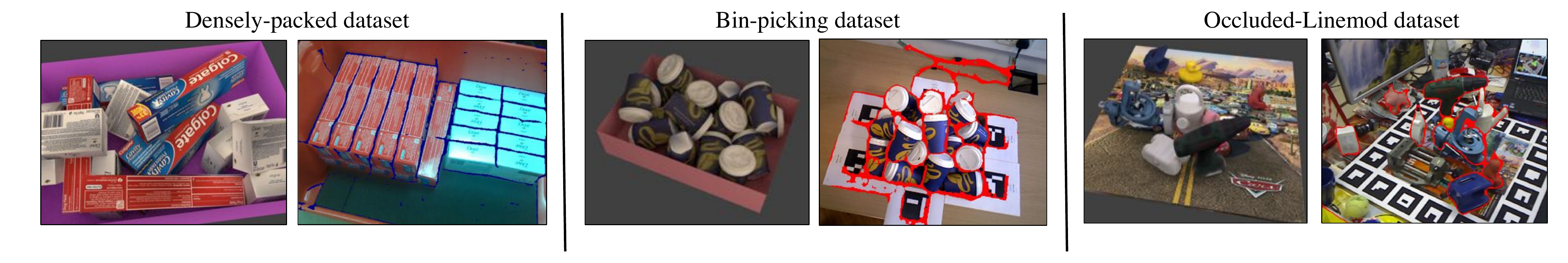}
\vspace{-0.2in}
\caption{Examples of synthetic training data and boundary predictions on real images.}
\vspace{-0.2in}
\label{fig:qualitative}
\end{figure}

Given the predictions from \cnn, the hypotheses generation process
described in Appendix A finds for each object type, a set of pose
candidates, which should be large enough to include the true poses and
small enough to reduce the computation time. The number of candidates
is affected by the number of sampled bases and pose clustering. Recall
rates are shown in Figure~\ref{hg_ablation} separately for all
generated candidates and the ones selected by {\tt ILP}. It also shows
the effect of dispersion parameter $\gamma$, which is by default set
to $0.9$. Table~\ref{hg_boundary} evaluates the contribution of the
boundary reasoning in the hypothesis generation and shows that it has
a significant impact on the overall recall.

Table~\ref{obj_function} compares learned objective function for {\tt
ILP} vs manually-defined ones. One way to combine features defined in
Section~\ref{sec:selection} is to consider only scene alignment scores
($f_3 + f_4$) as in many point registration algorithms or
alternatively use both scene alignment and model consistency scores
($f_1*f_2*f_3 (f_4 + f_5)$). An upper bound of performance with the
given hypotheses set is established as the optimal recall when the
true {\tt ADI} distance from ground-truth is used in the optimization.


The overall computation time for the current sequential implementation
of the approach ranges from 10s to 15s for estimating all (15 to 19)
instances in a scene. The \cnn\ predictions, the pose hypotheses
generation and the scene-level optimization along with collision
checking run individually in less than a second. Broad phase collision
checking is performed to speed up the process. The majority of
computation time is spent on depth-buffering and local refinement for
each pose candidate (130 per object category). These operations can be
significantly sped up with parallel processing such as {\tt
CUDA-OpenGL} interoperability, which has been shown \cite{sui2017goal}
to render 1000 images in 0.1s. Given the data parallelism, multiple
cores can also be easily utilized to speed up the algorithm.

\vspace{-.1in}

\section{Discussion}
\label{sec:conclusion}
\vspace{-.1in}

This work focuses on hard instances of scene-level, multi-instance
pose estimation, which includes highly cluttered and densely packed
scenarios. The results show that the type of poses used in simulation
for training the semantic and the boundary networks is important.
While a simulation-to-reality domain gap exists, it can be bridged by
using appropriate information, such as boundary prediction, which
translates well from simulated to real images, and adversarial
training strategies. Furthermore, the {\tt ILP} formulation of the
scene-level reasoning is able to find combinations of hypotheses that
are both consistent as well as of high-quality given the learned
global score function. The consistency of pose hypotheses in the
current work is defined by collision constraints. It is interesting to
consider a learning process for identifying compatible sets of poses
that express physical constraints and which could be used in the
context of the proposed {\tt ILP} formulation.


\acknowledgments{This work was supported by the NSF, grant numbers IIS-1734492 and IIS-1723869.}


\bibliography{root}  

\begin{thebibliography}{45}
\providecommand{\natexlab}[1]{#1}
\providecommand{\url}[1]{\texttt{#1}}
\expandafter\ifx\csname urlstyle\endcsname\relax
  \providecommand{\doi}[1]{doi: #1}\else
  \providecommand{\doi}{doi: \begingroup \urlstyle{rm}\Url}\fi

\bibitem[Correll et~al.(2016)Correll, Bekris, Berenson, Brock, Causo, Hauser,
  Osada, Rodriguez, Romano, and Wurman]{Correll:2016aa}
N.~Correll, K.~Bekris, D.~Berenson, O.~Brock, A.~Causo, K.~Hauser, K.~Osada,
  A.~Rodriguez, J.~Romano, and P.~Wurman.
\newblock {Analysis and Observations From the First Amazon Picking Challenge}.
\newblock \emph{T-ASE}, 2016.

\bibitem[Schwarz et~al.(2018)Schwarz, Lenz, Garc{\'\i}a, Koo, Periyasamy,
  Schreiber, and Behnke]{schwarz2018fast}
M.~Schwarz, C.~Lenz, G.~Garc{\'\i}a, S.~Koo, A.~Periyasamy, M.~Schreiber, and
  S.~Behnke.
\newblock Fast object learning and dual-arm coordination for cluttered stowing,
  picking, and packing.
\newblock In \emph{ICRA}, 2018.

\bibitem[Zeng et~al.()Zeng, Yu, Song, Suo, Walker, Rodriguez, and
  Xiao]{zeng2017multi}
A.~Zeng, K.~Yu, S.~Song, D.~Suo, E.~Walker, A.~Rodriguez, and J.~Xiao.
\newblock Multiview self-supervised deep learning for 6d pose estimation in the
  amazon picking challenge.
\newblock In \emph{ICRA'17}.

\bibitem[Gualtieri et~al.(2017)Gualtieri, Ten~Pas, Saenko, and
  Platt]{Gualtieri:2017aa}
M.~Gualtieri, A.~Ten~Pas, K.~Saenko, and R.~Platt.
\newblock {Grasp Pose Detection in Point Clouds}.
\newblock \emph{IJRR}, 2017.

\bibitem[Mahler et~al.(2017)Mahler, Liang, Niyaz, Laskey, Doan, Liu, Ojea, and
  Goldberg]{Mahler:2017aa}
J.~Mahler, J.~Liang, S.~Niyaz, M.~Laskey, R.~Doan, X.~Liu, J.~Ojea, and
  K.~Goldberg.
\newblock Dex-net 2.0.
\newblock In \emph{R:SS}, 2017.

\bibitem[Morrison et~al.(2018)]{morrison2018cartman}
D.~Morrison et~al.
\newblock Cartman: The low-cost cartesian manipulator that won the amazon
  robotics challenge.
\newblock In \emph{ICRA}, 2018.

\bibitem[Zeng et~al.(2018)]{zeng2018robotic}
A.~Zeng et~al.
\newblock Robotic pick-and-place of novel objects in clutter with
  multi-affordance grasping and cross-domain image matching.
\newblock In \emph{ICRA}, 2018.

\bibitem[Shome et~al.(2019)Shome, Tang, Song, Mitash, Kourtev, Yu, Boularias,
  and Bekris]{shome2019towards}
R.~Shome, W.~N. Tang, C.~Song, C.~Mitash, C.~Kourtev, J.~Yu, A.~Boularias, and
  K.~Bekris.
\newblock Towards robust product packing with a minimalistic end-effector.
\newblock In \emph{ICRA}, 2019.

\bibitem[Fan and Hauser(2019)]{Fan:2019aa}
W.~Fan and K.~Hauser.
\newblock {Robot Packing with Known Items and Nondeterministic Arrival Order}.
\newblock In \emph{R:SS}, 2019.

\bibitem[Hodan et~al.(2018)Hodan, Michel, Brachmann, Kehl, GlentBuch, Kraft,
  Drost, Vidal, Ihrke, Zabulis, et~al.]{hodan2018bop}
T.~Hodan, F.~Michel, E.~Brachmann, W.~Kehl, A.~GlentBuch, D.~Kraft, B.~Drost,
  J.~Vidal, S.~Ihrke, X.~Zabulis, et~al.
\newblock Bop: benchmark for 6d object pose estimation.
\newblock In \emph{ECCV}, 2018.

\bibitem[Aldoma et~al.(2012)]{aldoma2012tutorial}
A.~Aldoma et~al.
\newblock Tutorial: Point cloud library: Three-dimensional object recognition
  and 6 dof pose estimation.
\newblock \emph{IEEE Robotics \& Automation Magazine}, 2012.

\bibitem[Hinterstoisser et~al.(2012)Hinterstoisser, Lepetit, Ilic, Holzer,
  Bradski, Konolige, and Navab]{hinterstoisser2012model}
S.~Hinterstoisser, V.~Lepetit, S.~Ilic, S.~Holzer, G.~Bradski, K.~Konolige, and
  N.~Navab.
\newblock {Model based training, detection and pose estimation of texture-less
  3D objects in heavily cluttered scenes}.
\newblock In \emph{ACCV}, 2012.

\bibitem[Drost et~al.(2010)Drost, Ulrich, Navab, and Ilic]{drost2010model}
B.~Drost, M.~Ulrich, N.~Navab, and S.~Ilic.
\newblock {Model Globally, Match Locally: Efficient and Robust 3D Object
  Recognition}.
\newblock In \emph{CVPR}, 2010.

\bibitem[Vidal et~al.(2018)Vidal, Lin, Llad{\'o}, and Mart{\'\i}]{Vidal:2018aa}
J.~Vidal, C.-Y. Lin, X.~Llad{\'o}, and R.~Mart{\'\i}.
\newblock {A Method for 6D Pose Estimation of Free-Form Rigid Objects Using
  Point Pair Features on Range Data}.
\newblock \emph{Sensors}, 2018.

\bibitem[Doumanoglou et~al.(2016)Doumanoglou, Kouskouridas, Malassiotis, and
  Kim]{doumanoglou2016recovering}
A.~Doumanoglou, R.~Kouskouridas, S.~Malassiotis, and T.-K. Kim.
\newblock Recovering 6d object pose and predicting next-best-view in the crowd.
\newblock In \emph{CVPR}, 2016.

\bibitem[Xiang et~al.(2018)Xiang, Schmidt, Narayanan, and
  Fox]{xiang2017posecnn}
Y.~Xiang, T.~Schmidt, V.~Narayanan, and D.~Fox.
\newblock {PoseCNN: A Convolutional Neural Network for 6D Object Pose
  Estimation in Cluttered Scenes}.
\newblock In \emph{R:SS}, 2018.

\bibitem[Kehl et~al.(2017)Kehl, Manhardt, Tombari, Ilic, and
  Navab]{kehl2017ssd}
W.~Kehl, F.~Manhardt, F.~Tombari, S.~Ilic, and N.~Navab.
\newblock Ssd-6d: Making rgb-based 3d detection and 6d pose estimation great
  again.
\newblock In \emph{ICCV}, 2017.

\bibitem[Brachmann et~al.(2014)Brachmann, Krull, Michel, Gumhold, Shotton, and
  Rother]{brachmann2014learning}
E.~Brachmann, A.~Krull, F.~Michel, S.~Gumhold, J.~Shotton, and C.~Rother.
\newblock Learning 6d object pose estimation using 3d object coordinates.
\newblock In \emph{ECCV}, 2014.

\bibitem[Mitash et~al.(2018)Mitash, Boularias, and Bekris]{mitash2018robust}
C.~Mitash, A.~Boularias, and K.~Bekris.
\newblock Robust 6d object pose estimation with stochastic congruent sets.
\newblock In \emph{BMVC}, 2018.

\bibitem[Tremblay et~al.(2018)Tremblay, To, Sundaralingam, Xiang, Fox, and
  Birchfield]{Tremblay:2018aa}
J.~Tremblay, T.~To, B.~Sundaralingam, Y.~Xiang, D.~Fox, and S.~Birchfield.
\newblock {Deep Object Pose Estimation for Semantic Robotic Grasping of
  Household Objects}.
\newblock In \emph{CoRL}, 2018.

\bibitem[Sundermeyer et~al.(2018)Sundermeyer, Marton, Durner, Brucker, and
  Triebel]{Sundermeyer:2018aa}
M.~Sundermeyer, Z.-C. Marton, M.~Durner, M.~Brucker, and R.~Triebel.
\newblock {Implicit 3D Orientation Learning for 6D Object Detection from RGB
  Images}.
\newblock In \emph{ECCV}, 2018.

\bibitem[Aldoma et~al.(2012)Aldoma, Tombari, Di~Stefano, and
  Vincze]{aldoma2012global}
A.~Aldoma, F.~Tombari, L.~Di~Stefano, and M.~Vincze.
\newblock A global hypotheses verification method for 3d object recognition.
\newblock In \emph{ECCV}, 2012.

\bibitem[Elith et~al.()Elith, Leathwick, and Hastie]{elith2008working}
J.~Elith, J.~R. Leathwick, and T.~Hastie.
\newblock A working guide to boosted regression trees.
\newblock \emph{JAE'08}.

\bibitem[Narayanan and Likhachev(2016)]{Narayanan:2016aa}
V.~Narayanan and M.~Likhachev.
\newblock {Discriminatively-guided Deliberative Perceptinon for Pose Estimation
  of Multiple 3D Object Instances}.
\newblock In \emph{R:SS}, 2016.

\bibitem[Sui et~al.(2017)Sui, Xiang, Jenkins, and Desingh]{sui2017goal}
Z.~Sui, L.~Xiang, O.~C. Jenkins, and K.~Desingh.
\newblock Goal-directed robot manipulation through axiomatic scene estimation.
\newblock In \emph{IJRR}, 2017.

\bibitem[Mitash et~al.(2018)Mitash, Boularias, and Bekris]{mitash2018improving}
C.~Mitash, A.~Boularias, and K.~E. Bekris.
\newblock Improving 6d pose estimation of objects in clutter via physics-aware
  monte carlo tree search.
\newblock In \emph{ICRA}, 2018.

\bibitem[Long et~al.(2015)Long, Shelhamer, and Darrell]{long2015fully}
J.~Long, E.~Shelhamer, and T.~Darrell.
\newblock Fully convolutional networks for semantic segmentation.
\newblock In \emph{CVPR}, 2015.

\bibitem[Noh et~al.(2015)Noh, Hong, and Han]{noh2015learning}
H.~Noh, S.~Hong, and B.~Han.
\newblock Learning deconvolution network for semantic segmentation.
\newblock In \emph{ICCV}, 2015.

\bibitem[Yang et~al.(2016)Yang, Price, Cohen, Lee, and Yang]{yang2016object}
J.~Yang, B.~Price, S.~Cohen, H.~Lee, and M.-H. Yang.
\newblock Object contour detection with a fully convolutional encoder-decoder
  network.
\newblock In \emph{CVPR}, 2016.

\bibitem[Li et~al.(2017)Li, Qi, Dai, Ji, and Wei]{li2017fully}
Y.~Li, H.~Qi, J.~Dai, X.~Ji, and Y.~Wei.
\newblock Fully convolutional instance-aware semantic segmentation.
\newblock In \emph{CVPR}, 2017.

\bibitem[Kirillov et~al.(2017)Kirillov, Levinkov, Andres, Savchynskyy, and
  Rother]{kirillov2017instancecut}
A.~Kirillov, E.~Levinkov, B.~Andres, B.~Savchynskyy, and C.~Rother.
\newblock Instancecut: from edges to instances with multicut.
\newblock In \emph{CVPR}, 2017.

\bibitem[Mitash et~al.(2017)Mitash, Bekris, and Boularias]{mitash2017self}
C.~Mitash, K.~Bekris, and A.~Boularias.
\newblock A self-supervised learning system for object detection using physics
  simulation and multi-view pose estimation.
\newblock In \emph{IROS}, 2017.

\bibitem[Tobin et~al.(2017)Tobin, Fong, Ray, Schneider, Zaremba, and
  Abbeel]{tobin2017domain}
J.~Tobin, R.~Fong, A.~Ray, J.~Schneider, W.~Zaremba, and P.~Abbeel.
\newblock Domain randomization for transferring deep neural networks from
  simulation to the real world.
\newblock In \emph{IROS}, 2017.

\bibitem[Hinterstoisser et~al.(2018)Hinterstoisser, Lepetit, Wohlhart, and
  Konolige]{hinterstoisser2018pre}
S.~Hinterstoisser, V.~Lepetit, P.~Wohlhart, and K.~Konolige.
\newblock On pre-trained image features and synthetic images for deep learning.
\newblock In \emph{ECCV}, 2018.

\bibitem[Zhu et~al.(2017)Zhu, Park, Isola, and Efros]{CycleGAN2017}
J.-Y. Zhu, T.~Park, P.~Isola, and A.~A. Efros.
\newblock Unpaired image-to-image translation using cycle-consistent
  adversarial networkss.
\newblock In \emph{ICCV}, 2017.

\bibitem[Tsai et~al.(2018)Tsai, Hung, Schulter, Sohn, Yang, and
  Chandraker]{Tsai_adaptseg_2018}
Y.-H. Tsai, W.-C. Hung, S.~Schulter, K.~Sohn, M.-H. Yang, and M.~Chandraker.
\newblock Learning to adapt structured output space for semantic segmentation.
\newblock In \emph{CVPR}, 2018.

\bibitem[Wang et~al.(2019)Wang, Xu, Zhu, Mart{\'\i}n-Mart{\'\i}n, Lu, Fei-Fei,
  and Savarese]{wang2019densefusion}
C.~Wang, D.~Xu, Y.~Zhu, R.~Mart{\'\i}n-Mart{\'\i}n, C.~Lu, L.~Fei-Fei, and
  S.~Savarese.
\newblock Densefusion: 6d object pose estimation by iterative dense fusion.
\newblock In \emph{CVPR}, 2019.

\bibitem[Mellado et~al.(2014)Mellado, Aiger, and Mitra]{mellado2014super}
N.~Mellado, D.~Aiger, and N.~Mitra.
\newblock {Super4PCS: Fast Global Pointcloud Registration via Smart Indexing}.
\newblock In \emph{Computer Graphics Forum}, 2014.

\bibitem[Hastad(1996)]{Hastad96cliqueis}
J.~Hastad.
\newblock Clique is hard to approximate within $n^{1-\epsilon}$.
\newblock In \emph{Acta Mathematica}, 1996.

\bibitem[Krause and Golovin(2014)]{0001G14}
A.~Krause and D.~Golovin.
\newblock \emph{Submodular Function Maximization.}
\newblock Cambridge Press, 2014.

\bibitem[Pedregosa et~al.(2011)Pedregosa, Varoquaux, Gramfort, Michel, Thirion,
  Grisel, Blondel, Prettenhofer, Weiss, Dubourg, et~al.]{pedregosa2011scikit}
F.~Pedregosa, G.~Varoquaux, A.~Gramfort, V.~Michel, B.~Thirion, O.~Grisel,
  M.~Blondel, P.~Prettenhofer, R.~Weiss, V.~Dubourg, et~al.
\newblock Scikit-learn: Machine learning in python.
\newblock \emph{JMLR}, 2011.

\bibitem[Buch et~al.(2017)Buch, Kiforenko, and Kraft]{buch2017rotational}
A.~G. Buch, L.~Kiforenko, and D.~Kraft.
\newblock Rotational subgroup voting and pose clustering for robust 3d object
  recognition.
\newblock In \emph{ICCV}, 2017.

\bibitem[He et~al.(2017)He, Gkioxari, Doll{\'a}r, and Girshick]{he2017mask}
K.~He, G.~Gkioxari, P.~Doll{\'a}r, and R.~Girshick.
\newblock Mask r-cnn.
\newblock In \emph{ICCV}, 2017.

\bibitem[Tejani et~al.(2014)Tejani, Tang, Kouskouridas, and Kim]{Tejani:2014aa}
A.~Tejani, D.~Tang, R.~Kouskouridas, and T.~K. Kim.
\newblock {Latent-class Hough Forests for 3D Object Detection and Pose
  Estimation}.
\newblock In \emph{ECCV}, 2014.

\bibitem[To et~al.(2018)To, Tremblay, McKay, Yamaguchi, Leung, Balanon, Cheng,
  Hodge, and Birchfield]{to2018ndds}
T.~To, J.~Tremblay, D.~McKay, Y.~Yamaguchi, K.~Leung, A.~Balanon, J.~Cheng,
  W.~Hodge, and S.~Birchfield.
\newblock {NDDS}: {NVIDIA} deep learning dataset synthesizer, 2018.

\end{thebibliography}


\begin{thebibliography}{8}
\providecommand{\natexlab}[1]{#1}
\providecommand{\url}[1]{\texttt{#1}}
\expandafter\ifx\csname urlstyle\endcsname\relax
  \providecommand{\doi}[1]{doi: #1}\else
  \providecommand{\doi}{doi: \begingroup \urlstyle{rm}\Url}\fi

\bibitem[Brachmann et~al.(2014)Brachmann, Krull, Michel, Gumhold, Shotton, and
  Rother]{brachmann2014learning}
E.~Brachmann, A.~Krull, F.~Michel, S.~Gumhold, J.~Shotton, and C.~Rother.
\newblock Learning 6d object pose estimation using 3d object coordinates.
\newblock In \emph{ECCV}, 2014.

\bibitem[Drost et~al.(2010)Drost, Ulrich, Navab, and Ilic]{drost2010model}
B.~Drost, M.~Ulrich, N.~Navab, and S.~Ilic.
\newblock {Model Globally, Match Locally: Efficient and Robust 3D Object
  Recognition}.
\newblock In \emph{CVPR}, 2010.

\bibitem[Hinterstoisser et~al.(2016)Hinterstoisser, Lepetit, Rajkumar, and
  Konolige]{hinterstoisser2016going}
S.~Hinterstoisser, V.~Lepetit, N.~Rajkumar, and K.~Konolige.
\newblock Going further with point pair features.
\newblock In \emph{ECCV}, 2016.

\bibitem[Mitash et~al.(2018)Mitash, Boularias, and Bekris]{mitash2018robust}
C.~Mitash, A.~Boularias, and K.~Bekris.
\newblock Robust 6d object pose estimation with stochastic congruent sets.
\newblock In \emph{BMVC}, 2018.

\bibitem[Michel et~al.(2017)Michel, Kirillov, Brachmann, Krull, Gumhold,
  Savchynskyy, and Rother]{michel2017global}
F.~Michel, A.~Kirillov, E.~Brachmann, A.~Krull, S.~Gumhold, B.~Savchynskyy, and
  C.~Rother.
\newblock Global hypothesis generation for 6d object pose estimation.
\newblock In \emph{CVPR}, 2017.

\bibitem[Mellado et~al.(2014)Mellado, Aiger, and Mitra]{mellado2014super}
N.~Mellado, D.~Aiger, and N.~Mitra.
\newblock {Super4PCS: Fast Global Pointcloud Registration via Smart Indexing}.
\newblock In \emph{Computer Graphics Forum}, 2014.

\bibitem[Huang et~al.(2017)Huang, Kwok, and Zhou]{Huang:2017aa}
J.~Huang, T.-H. Kwok, and C.~Zhou.
\newblock {V4PCS: Volumetric 4PCS Algorithm for Global Registration}.
\newblock \emph{Journal of Mechanical Design}, 2017.

\bibitem[Aiger et~al.(2008)Aiger, Mitra, and Cohen-Or]{aiger20084}
D.~Aiger, N.~Mitra, and D.~Cohen-Or.
\newblock {4-points Congruent Sets for Robust Pairwise Surface Registration}.
\newblock In \emph{ACM Transactions on Graphics (TOG)}, 2008.

\end{thebibliography}


\end{document}


\section*{Appendix A: Pose Hypotheses Generation}
Given the output of semantic class predictions $\segnet$ and boundary prediction $\boundarynet$, this step aims to
generate a set of 6D pose hypotheses $\{\{\trans_{ij} \}_{j=1}^{H_i} \}_{i=1}^{K}$, where $H_i$ denotes the
number of pose hypotheses for each object model $M_i$ and is larger
than the number of the object's instances in the scene, i.e, $H_i \gg
N_i$. Similar to the overall objective of the problem setup, each
$T_{ij} = (t_{ij}, R_{ij})$ is a candidate translation $t_{ij} \in
R^3$ and rotation $R_{ij} \in SO(3)$ of some instance of object type
$\model_i$ in the camera's reference frame, for each of the $K$ object
types present in the scene.

The sets of pose hypotheses should be sufficiently large to ensure
they contain candidates close enough to the unknown true poses, i.e, $\forall i \in [1, K],\forall j \in [1,N_i], \exists\
T \in \{\trans_{ij'} \}_{j'=1}^{H_i} : \| \trans
- \trans^g_{ij}\|\leq \epsilon,$ where $\trans$
is a hypothesis pose for object $M_i$ and $\trans^g_{ij}$ is the
ground truth pose of the $j^{th}$ instance of object $M_i$.  Since the
ground truth poses are unknown during testing, one cannot guarantee
that the previous desired property will be satisfied unless the
candidates are densely sampled from $SE(3)$, which would make the
search computationally expensive.  Therefore, it is important to
sample the candidates in a manner that balances their diversity and
their similarity to the ground truth given the semantic class
predictions $\segnet$ and models $\{M_i\}_{i=1}^{K}$.

Pose hypotheses generation is often performed using RANSAC-like
techniques~\cite{brachmann2014learning} or via hough
voting~\cite{drost2010model}, or various distinctive geometric
features~\cite{hinterstoisser2016going,mitash2018robust,michel2017global},
which ensure that multiple points can be sampled from the same
object. These methods do not incorporate boundary information, which
is important in cases where the surfaces of multiple object instances
are aligned. The following explains the proposed pose hypotheses
generation process, which utilizes both semantic segmentation and
instance-boundaries.

\begin{wrapfigure}{r}{1.45in}
\vspace{-0.2in}
  \centering \includegraphics[width=1.45in]{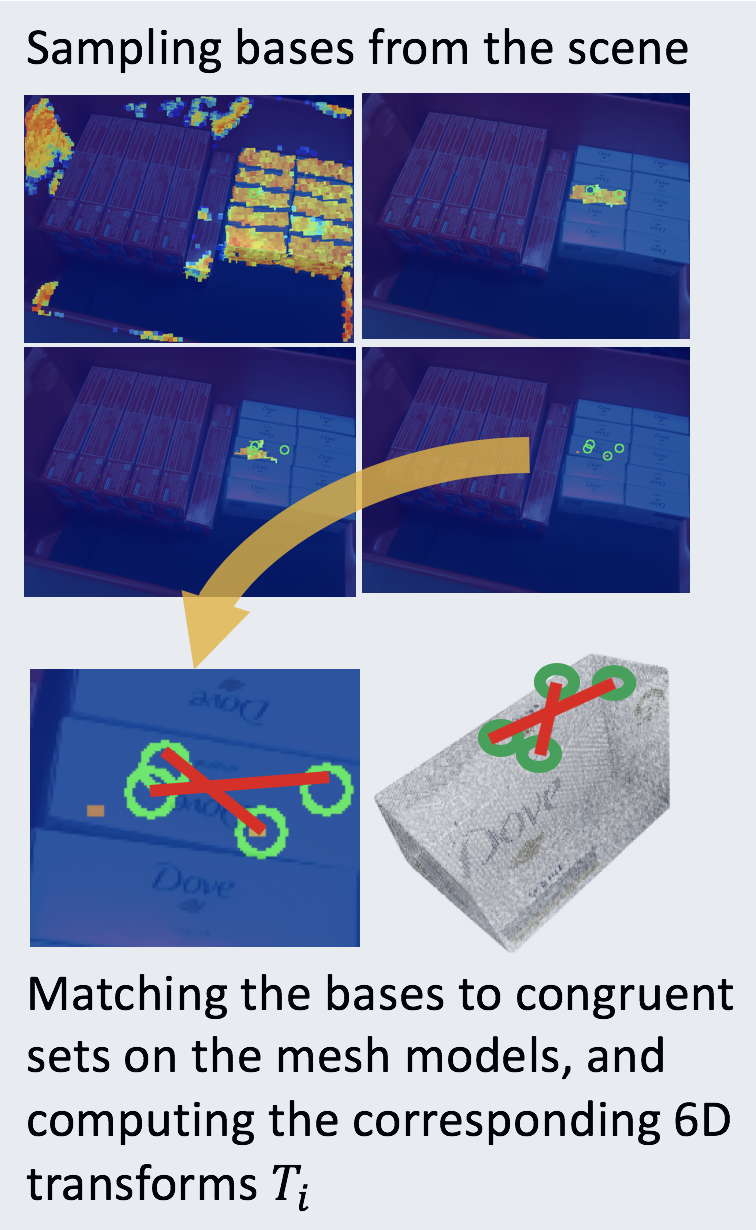}
\caption{Hypothesis generation process via sampling a base and
  matching it to a congruent set from the object model.}
  \label{fig:qualitative}
\vspace{-.2in}
\end{wrapfigure}

The output of the softmax layer from $\segnet$ is used to compute
$\pi_{l} (\pixel_i)$, the probability of each pixel $\pixel_i$
belonging to an object class $l$. This probability is defined as the
ratio of the output $\segnet(\pixel_i)$ over the sum of the outputs
for the same class over all pixels $\pixel$ in the given test image
$x$, i.e./ $\pi_{l} (\pixel_i)
= \frac{\segnet(\pixel_i)}{\sum_{\pixel}\segnet(\pixel)}.$

Based on the principles of randomized alignment, candidate poses are
generated for each instance of an object by sampling sets of points
that belong to one instance with high probability.  In the point cloud
registration literature \cite{mellado2014super, Huang:2017aa}, a set
of sampled points is called a {\it base} and denoted by $\base$.  Once
sampled, this set is matched to {\it congruent} sets, denoted by $U$,
on the corresponding object model $M$. Each of the congruent pairs
$(B,U)$ defines a unique rigid transform as long as both $B$ and $U$
contain at least 3 points each. For computational reasons, the
cardinality is typically limited to 4, i.e., $|\base| = |U| = 4$. To
maximize the probability that pose candidates for all the instances
are generated, a number $A_i$ of base sets $\{B_{ij} \}_{j=1}^{A_i}$
on the image $x$ is sampled, for each object class $i \in \{1, \dots,
K\}$. Each base $B_{ij}$ can be matched to several congruent bases $U$
on the models, and thus defines a large set of candidate poses.

\emph{A. Sampling a Base from a Single Object Instance:}
The four 3D points that form a base $\base = \{b_{1}, b_{2}, b_{3},
b_{4} \mid b_{1:4} \in S_l\}$ are sampled from a point cloud $S_l$,
which is the set of 3D points that are labeled as a category $l$
according to the semantic class probabilities $\segnet$.  The
challenge here is to maximize the joint probability of the four points
belonging to the same object instance. The joint probability
distribution on the graph formed by the four points is represented by
using the Hammersley-Clifford factorization and considering only unary
and binary relations between the points:

\begin{eqnarray}
	Pr\Big( \big( c(p(b_1),x) = l \big) \wedge \big(b_{1:4} \textrm{ are all on the same instance}\big)\Big)  \nonumber \\ = \frac{1}{Z}
\prod_{i=1}^{4} \Big( \potential^{l}_{node}(b_i) \prod_{j=1}^{j<i} \potential^{l}_{edge}(b_i,b_j) \Big), \nonumber
\end{eqnarray}

\noindent where $p(b_i)$ is used to denote the pixel corresponding to the 3D
point $b_i$ and, as before, $c(p,x)$ is used to denote the object
class label of pixel $p$ in image $x$. The unary terms
$\potential^{l}_{node}(b_i)$ are the individual probabilities of
labeling pixels $p(b_i)$ with category label $l$, i.e.,
$\potential^{l}_{node}(b_i) = \segnet(p(b_i))$. The binary relations
are defined as,

\begin{eqnarray*}
	\potential^{l}_{edge}(b_i, b_j) = 
	\begin{cases}
	1, & \text{ if } {\tt PPF}(b_i,b_j) \in {\tt PPF}(M_k) \text{ and} \cr
	   & \textrm{ \it \small length-of-shortest-path}(p(b_i),p(b_j)) < \epsilon_l \cr
	0, & \text{ otherwise}.
	\end{cases}
\end{eqnarray*}

The {\it Point-Pair Feature} ({\tt PPF}) \cite{drost2010model} for two
base points $b_i, b_j$ with surface normals $n_i, n_j$ is defined as:

\begin{align*}
	{\tt PPF}(b_i,b_j) = (\mid\mid d \mid\mid_2, \angle(n_i, d), \angle(n_j, d), \angle(n_i, n_j)),
\end{align*}

\noindent where $d=b_i-b_j$ is the vector from $b_i$ to $b_j$ and $n_i,
n_j$ are local surface normals.  ${\tt PPF}(M_k)$ is the set of
point-pair features pre-computed on the object model $M_k$ of object
category $k$. The \textrm {\it \small length-of-shortest-path} term is
the shortest distance on a graph $G$, where the vertices are the image
pixels and an edge connects two pixels $p_i$ and $p_j$ if and only if:
$P_B(p_i) < \delta$, $P_B(p_j) < \delta$, and $p_i$ and $p_j$ are
adjacent pixels in the image (each pixel has at most eight adjacent
pixels). In other terms, there is an edge between two adjacent pixels
if both pixels have a low probability (less than a threshold $\delta$)
of being on the boundary of an object. Thus, $\textrm{ \it \small
length-of-shortest-path}(p_i,p_j)$ is the length of the shortest path
between the two nodes on $G$, and it is obtained using a breadth-first
search.

\emph{B. Avoiding Oversampling and Achieving Coverage:}
The sampling process is dynamically adapted so as to avoid
repetitively selecting the same image regions. In particular, after a
base is sampled, a decay factor $\gamma\in[0,1)$ is multiplied to the
potential $\potential^{l}_{node}(b_i)$ of every point $b_i$ that
belongs to the same segment as $b_i$, i.e.,
$\potential^{l}_{node}(b_i)\leftarrow \gamma \potential^{l}_{node}(b_i)$. Points
belonging to the same segments are those encountered during the
breadth-first search. This encourages dispersion in the sampling
process of the bases, so as to cover all instances of objects in the
image and avoid having all the samples concentrated in the region of a
single object instance. This becomes an issue when a particular, more
prominent object instance has higher semantic class prediction
probability than other instances.


\emph{C. Matching the Base to its Congruent Sets:}
Once all the base sets are sampled, a matching process is used for
each one of the sampled bases $\{\{B_{ij} \}_{j=1}^{A_i} \}_{i=1}^{K}$
to compute a set $U_{ij}=\{u_1,u_2,u_3,u_4\}$ of 4-points from $M_i$
that are congruent to the sampled bases.  The basic idea of 4-point
congruent sets was derived from the fact that ratios and distances on
objects are invariant across any rigid
transformation~\cite{aiger20084}.

This matching process generates a large number of pose candidates
$\{\{\trans_{ij} \}_{j=1}^{H_i} \}_{i=1}^{K}$, several of which are
redundant due to different congruent pairs returning similar
transforms and the fact that most geometries have symmetry along
multiple axes. Thus, this work follows a greedy approach to provide a
diverse subset of the population. The approach sorts all pose
candidates based on the Largest Common Pointset
score~\cite{mellado2014super} and picks pose representatives that are
beyond a minimum distance from already selected poses. The distance
here takes into account symmetry, and there are separate thresholds
for rotation and translation, since these two variables operate in
different spaces. The resulting set
$\{\{\trans_{ij} \}_{j=1}^{H_i} \}_{i=1}^{K}$ of pose representatives
then undergoes local refinement using ICP.

\clearpage

\bibliography{supplement}  